\documentclass[11pt]{article}

\usepackage[final]{acl}
\usepackage{times}
\usepackage{latexsym}

\usepackage[T1]{fontenc}
\usepackage[utf8]{inputenc}
\usepackage[most]{tcolorbox}
\usepackage{microtype}

\usepackage{inconsolata}

\usepackage{graphicx}

\usepackage{fvextra}
\DefineVerbatimEnvironment{PromptBlock}{Verbatim}{
  breaklines=true,
  breakanywhere=true,
  fontsize=\footnotesize,
  breaksymbolleft={}
}
\usepackage[most]{tcolorbox}
\usepackage{listings}

\newtcolorbox{PromptBox}{
  colback=gray!4,
  colframe=black!60,
  boxrule=0.4pt,
  arc=2pt,
  left=6pt,
  right=6pt,
  top=6pt,
  bottom=6pt,
  breakable,
  enhanced,
  frame hidden=false,
  borderline={0pt}{0pt}{black!60},  % removes split line
}
\usepackage{tabularx}
\usepackage{stfloats}
\usepackage{booktabs}
\usepackage{tabularx}
\usepackage[table]{xcolor}

\usepackage[most]{tcolorbox}
\newtcolorbox{examplebox}{
  enhanced,
  breakable,
  colback=gray!5,
  colframe=black,
  boxrule=0.5pt,
  arc=2pt,
  left=4pt,
  right=4pt,
  top=4pt,
  bottom=4pt,
  before skip=8pt,
  after skip=8pt
}

\usepackage{alltt}
\usepackage{natbib}

\usepackage{cuted}
\usepackage{booktabs}
\usepackage{amsfonts}
\usepackage{subcaption}
\usepackage{placeins}

\PassOptionsToPackage{table}{xcolor}
\usepackage{xcolor}
\usepackage{multirow}
\usepackage{tabularx}
\usepackage{booktabs}

\newif\ifcomments
\commentsfalse

\ifcomments
  \newcommand{\aman}[1]{\textcolor{teal}{\textbf{[Aman: #1]}}}
  \newcommand{\eric}[1]{\textcolor{blue}{\textbf{[Eric: #1]}}}
  
\else
  \newcommand{\aman}[1]{}
  \newcommand{\eric}[1]{}
  \newcommand{\kai}[1]{}
\fi

\newcommand{\baselineOne}{single-dimensional}

\newcommand{\checklist}{Checklist-Style}
\newcommand{\flatrubrics}{Single-Dimensional}
\newcommand{\structured}{Multi-Dimensional}

\usepackage{dashrule}
\usepackage{xcolor}

\usepackage{tikz}
\usetikzlibrary{arrows.meta, positioning, fit, backgrounds, calc}

\newcounter{example}

\usepackage{enumitem}

\title{From Preferences to Principles: Rubric-Based\\Alignment for Grounded Knowledge Answers}

\author{
Aman Saini,  Priyanshu Kumar,  Eric Peng, Kai Yuan, Harsh Girase, Wanming Chen \\
Apple \\
\small \texttt{\{aman\_saini, priyanshu\_kumar, ericpeng, kai\_yuan, h\_girase, wanming\}@apple.com}
}

\begin{document}

\maketitle

\begin{abstract}
Designing effective reward signals for open-domain question answering is challenging because high-quality responses must simultaneously satisfy multiple aspects of answer quality that are difficult to capture with a holistic scalar objective. We introduce a rubric-based reward framework that generates query-specific rubrics grounded in retrieved evidence and decomposed into multiple quality dimensions, providing fine-grained supervision during post-training. Averaged across three evaluation axes (composition, grounding, and instruction-following), our approach improves over the instruction-tuned baseline by 6.5\% and over flat rubric variants by 4\%, with consistent gains across all evaluation datasets. Conditioning rubrics on retrieved evidence improves factual support, while decomposing rubrics into quality-specific dimensions further improves coherence, organization, and adherence to query requirements. Our results show that grounded, multi-dimensional rubrics provide more effective reward supervision for complex open-domain question answering.

\end{abstract}

\section{Introduction}

Reinforcement learning (RL) is widely used in language model post-training to optimize objectives beyond next-token prediction \cite{ouyang2022}. While RL is highly effective in domains with verifiable rewards such as coding and mathematics, extending it to open-ended tasks such as question answering and summarization remains challenging because answer quality is inherently multi-dimensional and difficult to capture through coarse preference signals \cite{gunjal2025, liang2023}. In practice, reward formulations based on opaque or weakly specified preference signals can lead to reward hacking \cite{gao2022scalinglawsrm, skalse2025definingcharacterizingrewardhacking}, verbosity inflation \cite{singhal2024longwaygoinvestigating, park2024disentanglinglengthqualitydirect}, and optimization toward superficial patterns \cite{sharma2025understandingsycophancylanguagemodels, casper2023openproblemsfundamentallimitations} rather than true answer quality. 
% add line break
Recent work has explored rubric-based and checklist-style supervision as mechanisms for representing explicit answer quality principles through structured feedback \cite{gunjal2025, viswanathan2025, zhang2026}. However, existing approaches typically rely on flat or weakly structured criteria and do not explicitly ground rubric generation in retrieved evidence. As a result, they provide limited control over how different dimensions of answer quality are independently represented and emphasized during training and may produce reward criteria that are insufficiently grounded in the available evidence. 
  To address this gap, we propose a rubric-based alignment framework that operationalizes explicit answer quality principles through query-specific, evidence-grounded, and multi-dimensional rubrics. This formulation enables fine-grained and controllable supervision for grounded question answering. 
 
 \noindent Accordingly, our contributions are as follows:

\begin{itemize}[itemsep=1pt, topsep=1pt, parsep=0pt]

\item We introduce a multi-dimensional rubric framework that organizes rubric criteria across distinct quality dimensions, enabling controllable supervision during post-training.

\item We propose retrieval-conditioned rubric generation using retrieved passages and reference answers to construct evidence-grounded supervision signals.

\item We empirically analyze retrieval conditioning and rubric decomposition for RL supervision, showing that the two components provide distinct and complementary gains.

\item We release the rubric generation and evaluation prompts used in our framework, along with the training query set\footnote{https://github.com/apple/ml-complex-qa-queries}, to support reproducibility and future research on rubric-based reward design.

\end{itemize}

\section{Related Work}

Recent post-training methods have shown that reinforcement learning is highly effective in domains with objective correctness signals, such as mathematical reasoning and code generation \cite{shao2024deepseekmath, le2022coderl}. However, extending these methods to open-ended language tasks such as question answering, instruction following, and summarization remains challenging because answer quality is subjective, multi-faceted, and difficult to capture with coarse rewards \cite{ouyang2022, liang2023}. Common approaches such as learned reward models and direct preference optimization \cite{rafailov2024dpo} rely on opaque preference signals that provide limited interpretability and are prone to spurious correlations.

A growing line of work addresses this challenge by replacing holistic reward supervision with structured natural-language criteria. 
% \textit{Rubrics as Rewards} \cite{gunjal2025} demonstrates that instance-specific rubric items can serve as reward signals for online reinforcement learning, while checklist-based approaches generate instruction-specific evaluation criteria and grade responses against these criteria using language models and verifier programs, producing supervision signals that can be used for preference optimization methods such as DPO \cite{viswanathan2025, zheng2023llmasjudge, rafailov2024dpo}. 
\textit{Rubrics as Rewards} \cite{gunjal2025} uses instance-specific rubrics as reward signals for online reinforcement learning, while checklist-based feedback (RLCF) \cite{viswanathan2025} extracts instruction-specific checklists and grades responses against them using AI judges and verifier programs. These approaches suggest that structured supervision provides a more informative and interpretable optimization signal than holistic preference judgments.

Recent work has further explored rubric design as a mechanism for improving reward robustness and scalability. \citet{zhang2026} show that rubric-based rewards can mitigate reward over-optimization \cite{gao2022scalinglawsrm} by better distinguishing between acceptable and high-quality outputs in the high-reward regime. Concurrently, OpenRubrics \cite{liu2026} and AdvancedIF \cite{he2025advancedif} study scalable rubric generation and rubric-based RL pipelines, shifting attention from coarse preference supervision toward systematic, structured supervision.

Despite these advances, prior work has largely focused on structured supervision in generic open-ended settings, with limited emphasis on retrieval-grounded rubric generation and controllable supervision across distinct quality dimensions. Our work addresses this gap through retrieval-conditioned and multi-dimensional rubric generation for retrieval-augmented generation (RAG) systems \cite{lewis2021rag}, where supervision must remain faithful to retrieved evidence.

% These limitations are particularly important for retrieval-augmented generation systems \cite{lewis2021rag} where supervision must remain faithful to retrieved evidence. Our work addresses these limitations through retrieval-conditioned and multi-dimensional rubric generation for grounded question answering.

\section{Problem Setup}

\subsection{Task Definition}

Our work focuses on grounded answer generation for knowledge-seeking queries in a RAG setting. Each example consists of a user query $q$ and a set of retrieved passages $D = \{d_1, \dots, d_n\}$, where $n$ is the number of passages. Given $(q, D)$, the policy is trained to generate a response $y$ that is helpful and answers the user query while being faithful to the provided evidence.

 Our setting differs from standard question answering in two important ways. First, the model is required to rely only on information supported by the retrieved passages, rather than on unstated external knowledge or potentially stale internal knowledge. Second, the generated answer must be explicitly attributable: every factual claim should be backed by at least one passage citation in the form \texttt{[index]}. This makes the objective more demanding than correctness alone, since a response may contain the correct answer while still being incomplete or insufficiently supported by the evidence.

 % Our work focuses on this setting, where the model must synthesize information across passages to produce grounded, well-structured responses. We study how to design reward signals that capture multiple dimensions of answer quality.

% We directly apply reinforcement learning to the instruction-tuned model without any additional supervised fine-tuning. We optimize the base policy using Group Relative Policy Optimization (GRPO) \cite{shao2024deepseekmath}.

% For each training example, we first generate instance-specific rubric criteria offline using the query, retrieved passages, and reference answer. The policy then generates multiple candidate answers conditioned on the query and retrieved passages. These responses are evaluated by an external LLM-based judge (Appendix~\ref{fig:judge-prompt-1}), which produces per-item judgments for the associated rubric criteria. The resulting judgments are aggregated into a scalar reward as described in Section~\ref{sec:reward-aggregation}, which GRPO uses to update the policy under a KL-regularized objective.

\subsection{Training Framework}
\begin{figure}[t]
\centering
\resizebox{\columnwidth}{!}{
\begin{tikzpicture}[
  font=\sffamily,
  box/.style={
    rounded corners=4pt,
    draw=black,
    thick,
    fill=white,
    minimum height=1.15cm,
    align=center,
    font=\sffamily\footnotesize,
    inner sep=4pt
  },
  llmbox/.style={
    rounded corners=4pt,
    draw=#1,
    thick,
    fill=#1!10,
    minimum height=1.15cm,
    align=center,
    font=\sffamily\footnotesize,
    inner sep=4pt
  },
  s2box/.style={
    rounded corners=4pt,
    draw=black,
    thick,
    fill=white,
    minimum width=2.45cm,
    minimum height=1.65cm,
    text width=2.05cm,
    align=center,
    font=\sffamily\footnotesize,
    inner sep=2pt
  },
  s2llmbox/.style={
    rounded corners=4pt,
    draw=green!50!black,
    thick,
    fill=green!50!black!10,
    minimum width=2.45cm,
    minimum height=1.65cm,
    text width=2.05cm,
    align=center,
    font=\sffamily\footnotesize,
    inner sep=2pt
  },
  arrow/.style={-{Latex[length=2mm]}, thick},
  title/.style={font=\sffamily\bfseries\small, anchor=west}
]

\def\stageW{11.6cm}
\def\stageH{2.75cm}

% ================= Stage 1 =================
\node[
  rounded corners=6pt,
  draw=blue,
  very thick,
  fill=blue!4,
  minimum width=\stageW,
  minimum height=\stageH
] (stage1) at (0,0) {};

\node[title, text=blue!60!black]
at ([xshift=-5.45cm,yshift=0.95cm]stage1.center)
{Stage 1: Rubric Generation (Offline)};

\node[box, text width=2.3cm] (s1in) at (-3.7,-0.25)
{Query +\\Passages +\\Reference Answer};

\node[llmbox=blue, text width=2.4cm] (s1llm) at (0,-0.25)
{{\bfseries\color{blue!80!black} Strong LLM\\Rubric Generator}\\[2pt]
{\scriptsize (\textit{Generation} prompt)}};

\node[box, text width=2.5cm] (s1out) at (3.7,-0.25)
{Multi-Dimensional\\Grounded Rubrics};

\draw[arrow] (s1in) -- (s1llm);
\draw[arrow] (s1llm) -- (s1out);

% ================= Stage 2 =================
\node[
  rounded corners=6pt,
  draw=green!50!black,
  very thick,
  fill=green!50!black!4,
  minimum width=\stageW,
  minimum height=3.25cm
] (stage2) at (0,-3.45) {};

\node[title, text=green!40!black]
at ([xshift=-5.45cm,yshift=1.18cm]stage2.center)
{Stage 2: RL with Rubric-based Rewards};

\node[s2box] (policy) at (-4.15,-3.45)
{Policy\\Model};

\node[s2box] (rollout) at (-1.4,-3.45)
{Rollout\\Response};

\node[s2llmbox] (s2judge) at (1.4,-3.45)
{{\bfseries\color{green!60!black} LLM Judge}\\[2pt]
{\scriptsize (\textit{Judge} Prompt)}};

\node[s2box] (reward) at (4.15,-3.45)
{Combined\\Reward\\$R = R_{\text{rubric}}$\\$-\lambda L$};

\draw[arrow] (policy.east) -- (rollout.west);
\draw[arrow] (rollout.east) -- (s2judge.west);
\draw[arrow] (s2judge.east) -- (reward.west);

% RL loop
\draw[-{Latex[length=2mm]}, thick, green!50!black]
  (reward.south) -- (4.15,-4.65)
  -- node[midway, below, font=\sffamily\scriptsize\bfseries]
     {RL Policy Update}
  (-4.15,-4.65)
  -- (policy.south);

% Rubric criteria flow into RL judge
\draw[
  dashed,
  -{Latex[length=2mm]},
  thick,
  blue!70!black
]
(s1out.south) ..
controls +(0,-1.45) and +(0,1.1) ..
node[pos=0.72, right, yshift=-3pt, font=\sffamily\scriptsize\bfseries, text=blue!70!black]
{Rubrics}
(s2judge.north);

% Stage arrow
\draw[arrow] (stage1.south) -- (stage2.north);

% % ================= Optional Evaluation =================
% \node[
%   rounded corners=6pt,
%   draw=orange,
%   very thick,
%   dashed,
%   fill=orange!4,
%   minimum width=\stageW,
%   minimum height=2.55cm
% ] (optional) at (0,-6.75) {};

% \node[title, text=orange!70!black]
% at ([xshift=-5.45cm,yshift=0.82cm]optional.center)
% {Optional: Rubric Quality Evaluation (Offline)};

% \node[box, text width=2.0cm] (optin) at (-4.15,-6.95)
% {Candidate\\Responses\\+ Rubrics};

% \node[llmbox=orange, text width=1.9cm] (optjudge) at (-1.4,-6.95)
% {{\bfseries\color{orange!80!black} LLM Judge}\\[2pt]
% {\scriptsize (\textit{Judge} Prompt)}};

% \node[box, text width=1.9cm] (optscore) at (1.4,-6.95)
% {Rubric Score\\$\in [0,1]$};

% \node[box, dashed, text width=2.0cm] (optsbs) at (4.15,-6.95)
% {SBS\\Preference\\Labels};

% \node[
%   font=\sffamily\scriptsize\itshape,
%   text=orange!70!black,
%   anchor=south
% ] at ([yshift=2pt]optsbs.north)
% {Alignment Validation};

% \draw[arrow] (optin.east) -- (optjudge.west);
% \draw[arrow] (optjudge.east) -- (optscore.west);
% \draw[<->, thick] (optscore.east) -- (optsbs.west);

\end{tikzpicture}
}
\caption{Overview of the proposed rubric generation and training pipeline. See Appendix~\ref{app:prompts} for prompt templates, particularly Figure~\ref{fig:rubric-prompt} for \textit{Generation} prompt and Figure~\ref{fig:judge-prompt-1} for \textit{Judge} prompt.}
\label{fig:rubric_pipeline}
\end{figure}
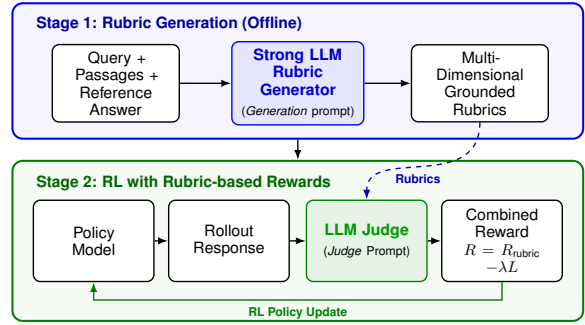

Figure~\ref{fig:rubric_pipeline} presents an overview of our rubric-based alignment pipeline. The core framework consists of two stages: offline rubric generation and online reinforcement learning with rubric rewards. 

We first generate instance-specific rubric criteria offline using the query, retrieved passages, and reference answer with a strong generative model. During training, the policy generates multiple candidate answers conditioned on the query and retrieved passages. These responses are evaluated by an external LLM-based reward judge \cite{zheng2023llmasjudge}, which produces per-item judgments for the associated rubric criteria. The resulting judgments are aggregated into a scalar reward, which is then used to optimize the policy using Group Relative Policy Optimization (GRPO) \cite{shao2024deepseekmath} under a KL-regularized objective.

% This setup cleanly separates \emph{generation} from \emph{evaluation}: the policy is optimized to produce grounded answers, while the judge provides structured supervision.

\section{Method}

We now describe our rubric-based reward formulation. We cover how query-specific rubrics are constructed, their structure and properties, and how per-item rubric judgments are aggregated into a final scalar reward for reinforcement learning.
                                                
\subsection{Rubric Generation}

% \begin{figure}[t]
%   \centering
%   \includegraphics[width=0.85\columnwidth]{Rubrics5.png}
%   \caption{Rubrics generation pipeline}
%   \label{fig:rubric_pipeline}
% \end{figure}

\subsubsection{Inputs}

For each query, we construct query-specific rubrics conditioned on three inputs using a rubric generation prompt (Figure~\ref{fig:rubric-prompt}): the user query $q$, a set of retrieved passages $D = \{d_1, \dots, d_n\}$, and a reference answer $\hat{y}$. Retrieved passages provide the factual grounding for rubric construction, while the reference answer serves as a high-quality example for capturing query intent and identifying crucial aspects of a good response. Importantly, the reference answer is not treated as ground truth, but as a guideline that helps identify the aspects a strong response should contain.

\subsubsection{Rubric Structure}

We organize rubrics into a fixed set of buckets, denoted by $\mathcal{B} = \{\texttt{accuracy},  \texttt{composition},  \texttt{safety},  \texttt{general}\}$. Together, these buckets operationalize the distinct principles of a high-quality response that guide reinforcement learning supervision. The \texttt{accuracy} bucket captures whether the response correctly addresses the user query and includes all necessary information to produce a complete answer. The \texttt{composition} bucket captures the structure, clarity, and presentation of the response. The \texttt{safety} bucket captures whether the response is safe and avoids harmful or misleading content. The \texttt{general} bucket captures factual consistency, groundedness, and domain-specific correctness not covered by other buckets. Within each bucket, we generate a set of rubric items, where each item specifies a single criterion for evaluating a candidate response. While we use this set of dimensions in our experiments, the formulation is not restricted to them and can be adapted to incorporate alternative or additional dimensions depending on task requirements.

\subsubsection{Rubric Properties}

For each query, we generate rubrics dynamically rather than relying on a fixed task-level specification. Our rubric generation framework (Figure~\ref{fig:rubric-prompt})  is guided by several principles intended to enable reliable criterion-level supervision during reward computation. Specifically, generated rubrics are designed to be query-specific, atomic, testable, evidence-grounded, and discriminative. 

Each rubric item is associated with a category (\texttt{Essential} or \texttt{Optional}) and a weight from 1 to 5, indicating its relative importance. \texttt{Essential} rubrics correspond to criteria whose failure would render a response fundamentally unhelpful, while \texttt{Optional} rubrics capture additional fine-grained aspects of response quality.

Examples of multi-dimensional rubrics generated for training queries, including bucket decomposition, essential/optional categorization, and rubric weighting, are provided in Appendix~\ref{app:example-rubric}.

\subsection{Reward Formulation}

During RL training, each query is paired with a set of precomputed rubrics, and the policy is optimized using a scalar reward derived from rubric satisfaction. An external LLM-based reward judge evaluates candidate responses against these rubrics using the judge prompt (Figure~\ref{fig:judge-prompt-1}), producing per-item binary judgments.

% This setup separates \emph{reward construction} from \emph{policy optimization}.

The reward judge evaluates responses using only the query, rubric, and generated answer, without direct access to retrieved passages. Grounding supervision is induced implicitly through rubric items generated from retrieved evidence during the offline rubric generation stage. This reward-stage judge is distinct from the evaluation-time grounding judge described in Section~\ref{metrics}, which directly evaluates whether generated statements are supported by retrieved passages.

% The reward computation itself remains fixed across different rubric formulations; variation arises from how rubric items are generated and organized across buckets.

\subsubsection{Bucket-level Scoring}

Let $\mathcal{B}$ denote the set of rubric buckets. In the multi-dimensional setting, $\mathcal{B} = \{\texttt{accuracy}, \texttt{composition}, \texttt{safety}, \texttt{general}\}$; in the \baselineOne\ setting, all rubric items are assigned to a single \texttt{general} bucket. For each bucket $b \in \mathcal{B}$, let $\mathcal{R}_b$ denote the set of rubric items in bucket $b$. Each rubric item $r \in \mathcal{R}_b$ is associated with a weight $w_r$ reflecting its relative importance.

\noindent Given a response $y$, the judge produces a binary satisfaction indicator $I_r(y) \in \{0,1\}$ for each rubric item $r$. The score for bucket $b$ is defined as
\[
S_b(y) = \sum_{r \in \mathcal{R}_b} w_r \, I_r(y)
\]

\noindent This scoring formulation applies uniformly across both rubric settings, isolating the effect of rubric design from reward computation.

\subsubsection{Reward Aggregation}
\label{sec:reward-aggregation}

We aggregate bucket-level scores to obtain a scalar reward. Let $\alpha_b$ denote the coefficient for bucket $b$, reflecting its relative importance, with $\sum_{b \in \mathcal{B}} \alpha_b = 1$. The weighted score for a response $y$ is given by
\begin{equation*}
T(y) = \sum_{b \in \mathcal{B}} \alpha_b S_b(y),
\end{equation*}
where $S_b(y)$ is the bucket-level score defined above. Because the number and weights of rubric items may vary across examples, we normalize by the maximum achievable weighted score for each example. The final reward is therefore
\begin{equation*}
R(y) =
\frac{\sum_{b \in \mathcal{B}} \alpha_b S_b(y)}
{\sum_{b \in \mathcal{B}} \alpha_b \sum_{r \in \mathcal{R}_b} w_r},
\label{eq:reward}
\end{equation*}

\noindent which ensures $R(y) \in [0,1]$.

This formulation enables explicit control over the optimization objective by adjusting the coefficients $\alpha_b$, allowing different aspects of answer quality to be prioritized without changing the reward architecture. We use this normalized reward to guide the policy toward responses that are not only correct but also grounded, complete, well-composed, and safe. The final scalar reward reflects aggregated satisfaction of multiple explicit and interpretable principles of answer quality, while also enabling fine-grained analysis of model behavior.

\section{Experiments}
\subsection{Training Setup}

We train our models using Group Relative Policy Optimization (GRPO) \cite{shao2024deepseekmath} implemented via the TRL\footnote{\url{https://github.com/huggingface/trl}} framework. We use \texttt{Qwen2.5-14B-Instruct} \citep{qwen2025qwen25}, an open-weight language model with 14.7B trainable parameters, released under the Apache 2.0 license, as the policy model. For each training prompt, we sample $K=4$ candidate completions from the current policy and optimize the model under a KL-regularized objective.

We train for 1 epoch with a per-device batch size of 8 and no gradient accumulation using 8 NVIDIA B200 GPUs. The learning rate is set to $1\times10^{-6}$ with a cosine learning rate schedule. For training-time candidate generation, we use a maximum completion length of 1024 tokens, temperature 0.5, and top-$p$ sampling with $p=0.9$. At evaluation time, we use greedy decoding (temperature 0) for deterministic outputs. Each run takes ${\sim}$24 GPU hours, totaling ${\sim}$240 GPU hours across all our experiments. Hyperparameters follow standard defaults and are held fixed across all experiments to isolate the effect of rubric design.

Rewards are computed using an external GPT-4o-based rubric judge with the aggregation logic defined in Section~\ref{sec:reward-aggregation}. We additionally apply a length-based penalty to discourage overly verbose responses, which is subtracted from the rubric-based reward as detailed in Appendix~\ref{sec:length-penalty}.

\subsection{Training Data}

We construct a synthetic training dataset of 4.7k knowledge-seeking queries in English designed to reflect realistic open-domain information needs in a RAG setting. Existing open-domain QA datasets often focus on short factoid-style questions and do not adequately capture the diversity and ambiguity of realistic assistant-style queries. In particular, many public datasets underrepresent phenomena such as underspecification, multi-intent behavior, entity and temporal ambiguity, and open-ended synthesis over retrieved evidence. We therefore synthetically generate training queries tailored to realistic question answering scenarios. 

We intentionally separate training and evaluation distributions in our experiments to evaluate generalization in open-domain question answering. Existing evaluation benchmarks are primarily designed for standardized assessment and either do not provide corresponding training splits or contain training distributions that do not adequately capture the diversity and complexity needed for generalization to unseen and challenging open-domain question answering tasks.

To better reflect real-world usage patterns, queries are generated using prompt templates targeting diverse characteristics, including multi-aspect reasoning, comparisons, ambiguity, tradeoff analysis, decision-making, and location- or time-dependent requests. The dataset additionally includes natural language queries with partial specifications, implicit context, and varied user intent. We release this synthetic training query set and make it publicly available to support future research on rubric-based supervision in open-domain question answering.

For each query, we use the Brave Search API\footnote{\label{fn:myref}\url{https://api.search.brave.com}, Web Search API v1.} to retrieve supporting passages, which provide the evidence used for rubric construction. Reference answers are generated using \texttt{GPT-4o} \cite{openai2024gpt4ocard} with the top-$k$ retrieved passages ($k=10$) as context, using the answer generation prompt (Figure~\ref{fig:generation-prompt-1}). These reference answers serve as grounded examples for rubric generation and are not treated as ground truth during training. 

\subsection{Evaluation Setup}

\subsubsection{Evaluation Datasets}
We benchmark our models on open-domain question answering datasets in a retrieval-augmented generation setting. To isolate generation quality from retrieval effects, we fix the retrieved passages for each query and evaluate only the model's ability to produce grounded, well-structured, and useful responses. Our evaluation suite consists of:

\begin{enumerate}
    \item \textbf{Search Arena} \citep{miroyan2025search}: We filter the Search-Arena v1 7k\footnote{\url{https://huggingface.co/datasets/lmarena-ai/search-arena-v1-7k}. Prompts released under CC BY 4.0.} dataset to retain English, single-turn samples with a clear winner and length between 10 and 500 characters. We deduplicate queries by exact query text and remove samples containing URLs or code snippets, resulting in 1.8k queries. For each query, we retrieve the top-$k$ ($k=10$) relevant passages using the Brave Search API\footnotemark[3] and fix them across all model evaluations.
    
    \item \textbf{RAGBench} \citep{friel2024ragbench}: We use the general knowledge subset of \textit{RAGBench}, with 1.7k query--document pairs from the test splits of MS MARCO \cite{bajaj2018msmarco}, HotpotQA \cite{yang2018hotpotqa}, HAGRID \cite{kamalloo2023hagrid}, and ExpertQA \cite{malaviya2024expertqa}. We report metrics on the concatenated test samples. A detailed dataset description is provided in Appendix~\ref{app:ragbench_datasets}.
    
    \item \textbf{FACTS Grounding Public} %\citep{cheng2025facts} evaluates LLMs' capability to generate relevant and accurate long responses with respect to given document context in a user request.
    \citep{cheng2025facts, factsdeepmind}\footnote{A benchmark developed by Google DeepMind and Google Research, released under CC BY 4.0.}: 
    We evaluate the model's ability to generate factually accurate responses grounded in provided document context using the 860 examples from the FACTS Grounding Benchmark \footnote{\url{https://huggingface.co/datasets/google/FACTS-grounding-public}}. Unlike standard QA benchmarks, FACTS requires models to synthesize information from long-form inputs and evaluates whether responses are fully supported by the given evidence.
\end{enumerate}

\subsubsection{Metrics} 
\label{metrics}

Given a model $f$ and an evaluation set consisting of query, retrieved content, and reference response tuples $\mathcal{D} = \{(q_i, r_i, \text{ref}_i)\}_{i=1}^{N}$, we use an LLM-as-a-judge $J$ to evaluate model generations along three axes: \textit{\textbf{Composition}, \textbf{Grounding}, and \textbf{Instruction-Following}}. To avoid evaluating the policy with the same model family that produced its training reward, we use \textbf{\textit{Gemini-2.5-Pro}}, a frontier LLM,  as the evaluation judge. The Composition evaluation prompt is custom-designed, while prompts for Grounding and Instruction-Following follow the FACTS Starter Benchmark\footnote{\url{https://www.kaggle.com/code/prathameshbang/facts-grounding-v2-benchmark-starter}}.

% We use \textit{Gemini-2.5-Pro} as the underlying state-of-the-art LLM judge. 

% \paragraph{\textbf{Composition}}  A given response to a query is well composed if it is 1) clear, 2) concise and direct, 3) structured and organized, 4) conversational, 5) useful, and 6) well presented. The LLM-as-a-judge generates a score between 1-5 for the above sub-axes and aggregates them into a final metric ranging from 1-5. We define the \textit{Composition} score of the model as:
% \[
%     Composition(f, D) = \frac{1}{N} \sum_{i=1}^{N} J(q_i, f(q_i, r_i))
% \]

\paragraph{\textbf{Composition}} Composition measures the presentation quality of a response, including clarity, conciseness, structure, completeness, tone, usefulness, and formatting. We use an LLM-as-a-judge prompt (Figure~\ref{fig:composition-prompt}) that evaluates each response along these seven dimensions on a 1--5 scale and aggregates them into a single score. We define the Composition score as:
\[
    \text{\textit{Comp}}(f, \mathcal{D}) = \frac{1}{N} \sum_{i=1}^{N} J_{\text{comp}}(q_i, f(q_i, r_i), \text{ref}_i),
\]
where $f(q_i, r_i)$ is the model-generated response to query $q_i$ given retrieved content $r_i$, $\text{ref}_i$ is the reference response, and $J_{\text{comp}}(\cdot)$ is the aggregated composition score.

% \paragraph{Grounding} A given response to a query is grounded to the retrieved documents if all information present in the response can be traced back to the retrieved content. We define the \textit{Grounding} score of a model as the percentage of grounded sentences. The LLM-as-judge module decomposes a response into sentences, which are then verified against the retrieved content; the judge categorizes each sentence as Supported, Not Supported, or NoRad (refers to sentences that do not require grounding in retrieved content). The \textit{Grounding} score is defined as:
% \begin{gather*}
% f(q, r) \rightarrow S = \{s_1, s_2, \ldots, s_n\} \\
% J(S, r) \rightarrow \{l_1, l_2, \ldots, l_n\} \\
% \ell_i \in \{\text{Supported},\ \text{NotSupported},\ \text{NoRad}\} \\
% Grounding(f, D) = \frac{1}{N_s} \sum_{i=1}^{N_s} \mathbf{1}\left[ l_i = \text{Supported}\right]
% \end{gather*}

% where $N_s$ is the total number of sentences accumulated across all $N$ query-retrieved content pairs in $D$

\paragraph{Grounding} A response is considered grounded when its claims are supported by the retrieved content. We measure grounding at the sentence level. Given a response, the LLM-as-a-judge decomposes it into sentences and classifies each sentence as \textit{Supported}, \textit{NotSupported}, or \textit{NoRad} (i.e., sentences that do not require grounding in the retrieved content). Let $f(q_i, r_i) \rightarrow S_i = \{s_{i1}, s_{i2}, \ldots, s_{i n_i}\}$ denote the set of sentences in the generated response. The judge assigns labels $J_{\text{ground}}(S_i, r_i) \rightarrow \{l_{i1}, l_{i2}, \ldots, l_{i n_i}\}$, where
 \[
l_{ij} \in \{\text{Supported},\ \text{NotSupported},\ \text{NoRad}\}
 \]

\noindent Let $\mathcal{L}_{\text{pass}} = \{\text{Supported}, \text{NoRad}\}$ denote labels that are counted as grounded. We define the Grounding score as:
\[
\text{\textit{Grounding}}(f, \mathcal{D}) =
\frac{1}{N_s}
\sum_{i=1}^{N} \sum_{j=1}^{n_i}
\mathbf{1}\left[l_{ij} \in \mathcal{L}_{\text{pass}}\right],
\]
where $N_s = \sum_{i=1}^{N} n_i$ is the total number of sentences across all responses.

% \paragraph{Instruction-Following} In a retrieval-augmented generation setting, Instruction-Following measures the helpfulness and usefulness of a response i.e., whether the response addresses the information requested by the query. $J$ takes as input the query, response, and a reference response to categorize the tuple as Major-Issue, Minor-Issue, or No-Issue. We define \textit{Instruction-Following (IF)} score as the fraction of responses without any Major-Issues:
% \begin{gather*}
% J(q_i, r_i, ref_i) = l_i \\
% l_i \in \{\text{MajorIssue}, \text{MinorIssue}, \text{NoIssue}\} \\
% IF(f, D) = 1 - \frac{1}{N} \sum_{i=1}^{N} \mathbf{1} \left[ l_i = \text{MajorIssue} \right] \\    
% \end{gather*}

\paragraph{Instruction-Following} This metric measures whether the response adequately addresses the information requested by the query. The LLM-as-a-judge takes as input the query, the model-generated response, and a reference response. It classifies the response into one of three categories: \textit{MajorIssue}, \textit{MinorIssue}, or \textit{NoIssue}. Formally, $J_{\text{IF}}(q_i, f(q_i, r_i), \text{ref}_i) = l_i$, where $l_i \in \{\text{MajorIssue}, \text{MinorIssue}, \text{NoIssue}\}$. We define the Instruction-Following (\textit{IF}) score as:
\[
\text{\textit{IF}}(f, \mathcal{D}) = 1 - \frac{1}{N} \sum_{i=1}^{N} \mathbf{1}\left[l_i = \text{MajorIssue}\right]
\]

% \noindent While the reward formulation and evaluation setup use different terminology, the underlying principles are closely related. The evaluation axes are not intended to exactly mirror the rubric bucket structure used during training. Instead, the rubric buckets define explicit quality dimensions used to steer and control model behavior during post-training, whereas evaluation uses externally established metrics whenever available.

\noindent Together, these axes assess different aspects of response quality. While the evaluation setup and reward formulation use different terminology, the underlying principles are closely related. The evaluation setup is not designed to mirror the rubric bucket structure used during training; instead, the rubric buckets define explicit quality dimensions used to steer and control model behavior during post-training, whereas evaluation uses externally established metrics whenever available.

\subsection{Baselines and Ablations}
We compare our approach against a strong instruction-tuned baseline and a set of ablations that progressively incorporate key components of our rubric-based reward formulation.

\paragraph{Instruction-Tuned Baseline:} We use the instruction-tuned Qwen2.5-14B-Instruct model\footnote{\url{https://huggingface.co/Qwen/Qwen2.5-14B-Instruct}} without additional fine-tuning as our baseline.

\paragraph{\checklist\ rubrics:} In this setting, we train using flat, checklist-style rubrics that are generated from the query and reference answer, without conditioning on retrieved content. All rubric items are placed in a single \texttt{general} bucket, without explicit decomposition into multiple dimensions during LLM structured decoding. This setting captures the effect of rubric-based supervision without grounding or structural decomposition, approximating the design of recent rubric-based approaches \citep{gunjal2025, viswanathan2025, zhang2026}.

\paragraph{\flatrubrics\ Grounded rubrics:}
 In this setting, rubrics are conditioned on retrieved content, but not decomposed into controllable quality dimensions. All rubric items are placed in a single \texttt{general} bucket, isolating the effect of grounding without multi-dimensional structure.

\paragraph{\structured\ Grounded rubrics:}
In our proposed setting, we include rubrics that are conditioned on retrieved content, along with multi-dimensional decomposition across predefined buckets (\texttt{accuracy}, \texttt{composition}, \texttt{safety}, and \texttt{general}). This setting enables both grounding and fine-grained control over different aspects of answer quality. Based on preliminary experiments, we assign higher weights to composition and accuracy as the primary drivers of answer quality, with safety and general acting as regularizing signals: $\alpha_{\texttt{composition}} = 0.50$, $\alpha_{\texttt{accuracy}} = 0.30$, and $\alpha_{\texttt{safety}} = \alpha_{\texttt{general}} = 0.10$.

\section{Results and Research Questions}
\label{sec:analysis}

% 1. Impact of Retrieval Conditioning
% 2. Impact of Structured output
% 3. Impact of Optional rubrics
% 4. Intrinsic evaluation of rubrics

\begin{table*}[t]
\centering
\small
\setlength{\tabcolsep}{3pt}
\begin{tabular}{lcccccccccc}
\toprule
& \multicolumn{3}{c}{Search Arena} 
& \multicolumn{3}{c}{RAGBench} 
& \multicolumn{3}{c}{FACTS} \\
\cmidrule(lr){2-4} \cmidrule(lr){5-7} \cmidrule(lr){8-10}
Model 
& \textit{Comp.} & \textit{Grounding} & \textit{IF}
& \textit{Comp.} & \textit{Grounding} & \textit{IF}  
& \textit{Comp.} & \textit{Grounding} & \textit{IF}  \\
\midrule

Qwen2.5-14B-Instruct 
& 3.65 & 81.80 & 73.80
& 4.37 & 82.50 & 94.80 
& 3.92 & 86.00 & 89.40 \\

\midrule

\checklist\ rubrics 
& 3.74 & 82.10 & 79.30
& 4.41 & 84.20 & 96.20 
& 4.00 & 87.90 & 91.20 \\

\flatrubrics\ Grounded rubrics 
& 3.71 & 83.20 & 79.10 
& 4.40 & 85.30 & 94.90 
& 3.98 & 88.00 & 89.50 \\

\structured\ Un-Grounded rubrics 
& 3.96 & 81.30 & 79.50 
& 4.49 & 84.20 & 96.10 
& 4.24 & 87.60 & \textbf{93.70} \\

\structured\ Grounded rubrics 
& \textbf{4.15} & \textbf{84.50} & \textbf{82.30} 
& \textbf{4.58} & \textbf{86.00} & \textbf{96.30}
& \textbf{4.39} & \textbf{88.60} & 91.50 \\

\bottomrule
\end{tabular}
\caption{
Main results across our three evaluation datasets. \textit{Comp.} refers to \textit{Composition} and is reported on a 1--5 scale. \textit{IF} refers to \textit{Instruction-Following}. \textit{Grounding} and \textit{IF} are reported as percentages. \textbf{Takeaway}: \textit{Model trained with grounded and structurally decomposed rubrics achieves the best overall performance.} 
}
\label{tab:main-results}
\end{table*}

Table~\ref{tab:main-results} presents the performance of the instruction-tuned baseline and all rubric-based variants across our evaluation suite. We observe that rubric-based training consistently improves over the baseline, with the largest gains achieved by combining retrieval conditioning and multi-dimensional rubric decomposition. Improvements are most pronounced on Search Arena, which requires complex, multi-aspect synthesis. Gains on RAGBench are smaller in absolute terms, given the already-high baseline performance. We perform a detailed analysis of the individual components, including retrieval conditioning, structured decomposition, and optional rubrics, in the following subsections.

\subsection{Do Retrieval-Conditioned Rubrics Improve Answer Quality?}

We observe that conditioning rubrics on retrieved evidence improves \textit{Grounding} across all datasets (Table~\ref{tab:main-results}). Both the \flatrubrics\ and \structured\ Grounded variants outperform the \checklist\ approach on \textit{Grounding}, indicating that retrieval-conditioned rubric rewards lead to stronger factual support and more consistent evidence attribution in generated responses.

Qualitative inspection in Appendix~\ref{app:qualitative-examples} further shows that grounded variants more consistently align claims with supporting citations and incorporate additional evidence-backed details compared to the baseline and checklist-style models, thus corroborating the above observation.

\subsection{Do Multi-Dimensional Rubrics Improve Answer Quality?}
%\subsection{Effect of Rubric Structuring on Answer Quality}

% We study the performance of the \structured\ grounded rubrics model, which has grounded rubrics decomposed into answer quality buckets. \textbf{We observe that it achieves the best performance across all metrics for all evaluation datasets} (Table \ref{tab:main-results}). Compared to the \flatrubrics\ grounded variant, it provides consistent improvements, indicating that decomposing rubrics into multiple dimensions of answer quality leads to more effective reward signals. This suggests that structured supervision enables more effective optimization by providing fine-grained signals across different aspects of answer quality.

We observe that rubric decomposition into answer quality buckets achieves the best performance across all metrics and datasets (Table~\ref{tab:main-results}). Compared to the \flatrubrics\ Grounded variant, the \structured\ Grounded model provides consistent improvements, with the largest gains on \textit{Composition} and \textit{Instruction-Following}, indicating that decomposing rubrics into quality-aligned dimensions provides fine-grained supervision for response coherence, completeness, synthesis, and adherence to query requirements.

Qualitative inspection in Appendix~\ref{app:qualitative-examples} shows that while both grounded variants provide strong factual support, the \structured\ Grounded model produces responses that are more coherent, better synthesized, and more effective at aggregating and presenting information. These findings suggest that structured rubric decomposition enables more effective reward optimization across complementary aspects of answer quality that holistic preference signals inherently fail to capture.

% \section{Analysis}
% \label{sec:analysis}

% \paragraph{Effect of Grounding and Decomposition.}
% Our results show that grounding and structural decomposition play distinct roles in improving answer quality. Introducing grounding through retrieval-conditioned rubrics improves factual support, as reflected in higher grounding scores. However, when used in isolation, grounding does not consistently improve overall answer quality.

%\paragraph{Effect of Rubric Structure.}
% We compare unstructured rubric formulations against our structured bucket-based approach. Structured rubrics consistently improve composition and instruction-following success, suggesting that decomposing quality into explicit dimensions leads to more stable and informative rewards.

%\paragraph{Effect of Rubric Conditioning.}
% We study the impact of conditioning rubric generation on retrieved passages and reference answers. Conditioning improves performance across datasets, indicating that incorporating evidence and strong target responses leads to more aligned supervision.

\subsection{Do Optional Rubrics Improve Answer Quality?}
%\subsection{Effect of Optional Rubrics on Answer Quality}
\begin{table*}[t]
\centering
\small
\setlength{\tabcolsep}{4pt}
\begin{tabular}{lccccccccc}
\toprule
& \multicolumn{3}{c}{Search Arena} 
& \multicolumn{3}{c}{RAGBench} 
& \multicolumn{3}{c}{FACTS} \\
\cmidrule(lr){2-4} \cmidrule(lr){5-7} \cmidrule(lr){8-10}
Model 
& \textit{Comp.} & \textit{Grounding} & \textit{IF} 
& \textit{Comp.} & \textit{Grounding} & \textit{IF} 
& \textit{Comp.} & \textit{Grounding} & \textit{IF} \\
\midrule

\texttt{Essential}  
& 3.87 & 83.60 & 79.00 
& 4.48 & 85.60 & 96.00 
& 4.21 & 87.20 & 90.00 \\

\texttt{Essential} + \texttt{Optional}
& \textbf{4.15} & \textbf{84.50} & \textbf{82.30} 
& \textbf{4.58} & \textbf{86.00} & \textbf{96.30} 
& \textbf{4.39} & \textbf{88.60} & \textbf{91.50} \\

\bottomrule
\end{tabular}
\caption{
Ablation comparing \texttt{Essential}  and \texttt{Essential} + \texttt{Optional} rubric formulations. \textit{Comp.} refers to \textit{Composition} and is reported on a 1--5 scale. \textit{IF} refers to \textit{Instruction-Following}. \textit{Grounding} and \textit{IF} are reported as percentages. \textbf{Takeaway}: \textit{Optional rubrics provide fine-grained reward signals, leading to higher-quality responses.}
}
\label{tab:ablation-optional}
\end{table*}

% During rubrics generation, each rubric is categorized as \texttt{essential} or \texttt{optional}. We compare a model trained using only \texttt{essential} rubrics against the model trained with \structured\  grounded rubrics, which includes both essential and optional rubrics. Both variants use retrieval-conditioned, multi-dimensional rubrics with identical bucket weights; the only difference is the inclusion of optional rubric items in the final reward aggregation. As shown in Table~\ref{tab:ablation-optional}, \textbf{incorporating optional rubrics improves performance across all metrics and datasets}, implying that optional criteria provide additional supervision to help the model better distinguish between partially correct and high-quality responses. While essential rubrics capture the minimum requirements for a valid answer, optional rubrics introduce finer-grained preferences that guide the model toward more complete and well-structured outputs. This highlights the importance of incorporating both critical and nuanced reward signals in rubric-based reward design.

We observe that incorporating \texttt{Optional} rubrics improves performance across all metrics and datasets, as shown in Table~\ref{tab:ablation-optional}. We compare a model trained using only \texttt{Essential} rubrics against the \structured\ Grounded model, which includes both \texttt{Essential} and \texttt{Optional} rubric items. Both variants use retrieval-conditioned, multi-dimensional rubrics with identical bucket weights; the only difference being the inclusion of \texttt{Optional} rubric items in the final reward aggregation. 

While \texttt{Essential} rubrics capture the minimum requirements for a valid answer, \texttt{Optional} rubrics introduce fine-grained preferences that help the model distinguish between satisfactory and high-quality responses. This highlights the importance of incorporating both critical and nuanced reward signals in rubric-based reward design.

\paragraph{Key Takeaways.}
Overall, the results show that both retrieval conditioning and multi-dimensional decomposition contribute to performance gains, with their combination yielding the strongest results. These findings highlight that moving beyond holistic reward signals toward structured, principle-based supervision is essential for aligning large language models on complex open-domain tasks.

% -------------------- Future sections --------------------

% \subsection{Rubric Quality Analysis}
% We analyze the quality of generated rubrics using an offline evaluation pipeline.
% Specifically, we measure whether rubric items are (i) relevant to the query,
% (ii) grounded in retrieved evidence, and (iii) aligned with high-quality
% reference responses. We find that rubric quality strongly correlates with
% downstream model performance, indicating that improvements in rubric
% generation translate directly into better training signals.

% \subsection{Case Studies}
% We present qualitative examples comparing model outputs across different
% training configurations. These examples highlight that rubric-based training
% leads to responses that are more structured, better grounded, and easier to
% verify.

% \subsection{Failure Cases}
% Despite improvements, we observe several failure modes. Models occasionally
% overfit to rubric phrasing, leading to overly verbose or templated responses.
% Grounding errors can persist when rubrics fail to capture subtle contradictions
% in retrieved passages. Finally, rubric quality may degrade for ambiguous queries,
% limiting the effectiveness of the reward signal.

% \input{latex/06_results}
% \input{latex/07_analysis}

\section{Rubric Quality Analysis}
\label{sec:rubric-quality}

The main evaluation (Table~\ref{tab:main-results}) measures end-to-end RL outcomes but does not isolate whether the generated rubrics align with unseen preference judgments. To validate rubric design choices independently of policy training, we compare rubric-derived preferences against LLM-based preference labels on the Question-Answering subset of OpenGenAlign \cite{opengenalign}. Additional dataset details are provided in Appendix~\ref{app:opengenalign}. Each instance consists of a query, retrieved passages, and two candidate responses from a pool of 12 LLMs. We generate rubrics under different configurations and compare the resulting preferences across three metrics:
\begin{enumerate}
    \item \textbf{Ties}: the fraction of pairs where both responses receive the same rubric score; lower values indicate more discriminative rubrics.
    \item \textbf{Agreement}: the fraction of pairs where the rubric-derived preference matches the preference label.
    \item \textbf{Egregious Errors}:
    the fraction of pairs where the rubric and the preference label select opposite responses (excluding ties).
\end{enumerate}

\begin{table}[t]
\centering
\small
\setlength{\tabcolsep}{4pt}
\begin{tabular}{lccc}
\toprule
Configuration & Ties & Agreement & Eg. Errors \\
\midrule
\checklist\ rubrics & 34.0 & 43.4 & 22.6 \\
+ Retrieval Conditioning & 27.8 & \textbf{54.0} & \textbf{18.2} \\
+ Decomposition & \textbf{23.8} & 53.4 & 20.8 \\
\bottomrule
\end{tabular}
% \caption{
% Preference alignment on OpenGenAlign. All values are percentages. \textbf{Takeaway}: Retrieval conditioning substantially improves preference alignment, while multi-dimensional decomposition further improves discriminativeness.
% }
\caption{
Preference alignment on OpenGenAlign. Higher is better for Agreement, while lower is better for Ties and Egregious Errors. 
% \textbf{Takeaway}: Retrieval conditioning substantially improves preference alignment, while multi-dimensional decomposition further improves discriminativeness.
}
\label{tab:rubric-quality}
\end{table}

\noindent Table~\ref{tab:rubric-quality} reports preference alignment under successive rubric design changes. \checklist\ rubrics achieve 43.4\% agreement with LLM-based preferences, suggesting that unstructured criteria provide weak signal for reliable preference estimation. Retrieval conditioning yields the largest improvement, increasing agreement by 10.6\% (43.4\% $\rightarrow$ 54.0\%) while reducing egregious errors from 22.6\% to 18.2\%. Rubric decomposition further reduces ties (27.8\% $\rightarrow$ 23.8\%), producing more discriminative rubric-derived preferences while maintaining comparable agreement. We hypothesize that this increased discriminativeness benefits GRPO training, since tied rewards between completions produce zero gradient signal. While we do not directly verify this mechanism, the observed pattern is consistent with the downstream RL gains in Table~\ref{tab:main-results}.
\section{Conclusion}

% We explore rubrics-based reinforcement learning for open-domain question answering by incorporating retrieval conditioning and structured decomposition during rubrics generation. We observe that retrieval conditioning of rubrics lead to more faithful responses, structured decomposition helps in improving composition and instruction-following metrics, and that optionally important rubrics impart finer reward signals during training.

% We present a rubric-based reinforcement learning framework for open-domain question answering that incorporates retrieval conditioning and structured decomposition. We observe that retrieval-conditioned rubrics lead to more faithful responses, while structured decomposition improves composition and instruction-following. Through quantitative and qualitative analysis, we demonstrate that structured reward design provides more effective supervision than flat reward formulations. These findings highlight the importance of decomposing reward signals for aligning language models in complex question answering tasks.

We present a rubric-based reinforcement learning framework for open-domain question answering that incorporates retrieval conditioning and structured decomposition. We show that retrieval-conditioned rubrics improve factual grounding and response faithfulness, while structured decomposition further improves composition and instruction-following. Through both quantitative and qualitative analysis, we demonstrate that moving from flat reward formulations toward structured, principle-based reward design provides a more effective foundation for aligning language models in complex question answering tasks.
\section{Limitations}

We describe several limitations of our work. First, we evaluate our approach on open-domain question answering, and its effectiveness in other domains and in multilingual settings remains to be validated. Second, response quality is inherently subjective, and our LLM-based evaluation serves as a proxy for human judgment. While recent work has shown strong correlation between frontier LLM judges and human preferences, more comprehensive human evaluation could further strengthen the conclusions. Third, due to compute constraints, all reported results correspond to single training and evaluation runs. While trends are consistent across three datasets and multiple ablations, small absolute differences may reflect run-to-run variance. Fourth, the quality of generated rubrics depends on the quality, diversity, and coverage of the reference answers used during rubric generation. In our setup, each query is associated with a single reference answer, which may introduce stylistic or content biases. Ideally, rubrics should be generated from multiple diverse reference answers from different model families, to capture common properties of high-quality responses rather than idiosyncrasies of any one generator. Finally, our framework assumes access to reasonably relevant retrieved evidence during both rubric generation and reference answer generation. Retrieval failures or incomplete evidence may reduce the quality of generated rubrics and downstream supervision signals, and our fixed retrieval evaluation setting does not fully capture retrieval errors in end-to-end systems.

\section{Ethical Considerations}

We discuss several ethical considerations related to rubric-based post-training for open-domain question answering systems. First, our framework relies on LLM-generated rubrics, reference answers, and evaluation judgments, which may inherit biases or inaccuracies from the underlying models. Second, LLM-generated synthetic data may introduce distributional biases or fail to capture the full diversity of real-world user behavior. Third, rubric-based optimization introduces the risk of over-optimization toward rubric satisfaction rather than broader notions of helpfulness or truthfulness. Although retrieval grounding can help reduce unsupported or hallucinated content, imperfect retrieval, rubric design, or alignment principles may still incentivize undesirable behaviors or factual errors. Human oversight and careful evaluation remain important for real-world deployment.

\bibliography{custom}

% Prompt templates in appendix- since they are long, need to split into separate entries
\appendix

\section{Additional Training Details}
\label{app:training-details}

\subsection{Length Penalty}
\label{sec:length-penalty}

We apply a length-based penalty to discourage overly verbose responses. This serves as a regularization term that mitigates verbosity-driven reward inflation while preserving the underlying rubric-based supervision. Intuitively, the penalty depends on the ratio between the generated and reference answer lengths. Responses that are close in length to the reference are not penalized, moderately longer responses incur a small penalty, and excessively long responses incur a larger penalty that is capped. Thresholds and coefficients were selected empirically based on preliminary experiments.

\noindent Formally, let $L_c$ and $L_r$ denote the number of sentences in the generated completion and reference answer, respectively. We define the length ratio as:
\begin{equation*}
r = \frac{L_c}{L_r}
\end{equation*}

\noindent The penalty is defined as a piecewise function of $r$:
\begin{equation*}
\begin{aligned}
\text{penalty}(r) ={}&
\begin{cases}
0, & r \leq 1.4, \\
0.1(r - 1.4), & 1.4 < r \leq 1.8, \\
0.04 + 0.08e, & r > 1.8,
\end{cases}
\\
&\text{where } e = \min(r - 1.8, 1.0).
\end{aligned}
\end{equation*}

\noindent The final reward is computed by subtracting this penalty from the normalized rubric-based reward $R(y)$ explained in Section~\ref{sec:reward-aggregation}:
\begin{equation*}
R'(y) = \max\big(0, \, R(y) - \text{penalty}(r)\big).
\end{equation*}

\section{Additional Details on Datasets}
\label{app:eval_datasets}

\subsection{RAGBench}
\label{app:ragbench_datasets}

We evaluate on datasets from the general knowledge subset of \textit{RAGBench}~\citep{friel2024ragbench}, which adapts standard open-domain question answering benchmarks into a retrieval-augmented generation (RAG) setting using query--document pairs.

\paragraph{Dataset selection.}
We pick MS MARCO (423), HotpotQA (390), HAGRID (716), and ExpertQA (203) datasets, resulting in a combined evaluation subset of 1.7k examples. These datasets are selected to cover a diverse range of query distributions, reasoning requirements, and answer formats commonly encountered in open-domain QA.

\paragraph{Dataset characteristics.}
\begin{itemize}
    \item \textbf{MS MARCO} \cite{bajaj2018msmarco}: Contains real-world user queries from web search logs. Questions are typically single-hop and require short, factoid-style answers grounded in relatively short passages. Released under the CC BY 4.0 license.
    
    \item \textbf{HotpotQA} \cite{yang2018hotpotqa}: A crowd-sourced dataset designed for multi-hop reasoning, where answering a query requires combining information from multiple documents or evidence chains. Released under the CC BY-SA 4.0 license.
    
    \item \textbf{HAGRID} \cite{kamalloo2023hagrid}: Focuses on grounded generation and hallucination robustness, with queries that require careful use of provided context to avoid unsupported claims. Released under the Apache 2.0 license.
    
    \item \textbf{ExpertQA} \cite{malaviya2024expertqa}: Consists of expert-authored questions that often require deeper domain knowledge, synthesis across longer contexts, and more detailed, long-form answers. Released under the MIT license.
\end{itemize}

\noindent These datasets differ not only in reasoning complexity (single-hop vs. multi-hop), but also in question source (real user queries, crowd-sourced, and expert-authored) and expected answer structure (short factual responses vs. long-form explanations). The datasets exhibit significant variation in both context length and answer complexity. For example, MS MARCO typically involves shorter passages and concise answers, whereas ExpertQA includes longer contexts and requires more detailed responses. This variation enables evaluation across both short-form and long-form generation regimes.

 Each example consists of a query paired with retrieved supporting documents. We use these query--document pairs as provided and do not modify the retrieval component. This allows us to isolate the quality of generation conditioned on a fixed set of retrieved passages. We use the test splits of all datasets to ensure evaluation on unseen data, and to minimize the risk of contamination from base model pretraining. The combination of datasets allows us to assess performance across a spectrum of query types, while also verifying that training on more complex queries does not regress performance on standard question answering tasks.

\subsection{OpenGenAlign}
\label{app:opengenalign}

We use 500 test samples from the Question-Answering subset of OpenGenAlign \cite{opengenalign} (released under the MIT license), which constructs pairwise preference labels over the WebGLM dataset \cite{liu2023webglm}. Each instance consists of a query, retrieved reference passages, and two candidate responses sampled from a pool of 12 LLMs.

 Preference labels are generated using o3 \cite{o3openai} through majority voting over three independent judgments. The OpenGenAlign authors validate these labels against human annotations, reporting 77\% agreement on the WebGLM subset \cite{opengenalign}. We use these labels to evaluate whether rubric-derived preferences align with LLM-based preference judgments under different rubric configurations.

\section{Examples of Multi-Dimensional Rubrics}
\label{app:example-rubric}
To illustrate the structure of our rubric generation framework, we present two training queries along with their multi-dimensional grounded rubrics. The examples demonstrate how rubric items are decomposed across quality dimensions, categorized as \texttt{Essential} or \texttt{Optional}, and weighted from 1--5 according to their relative importance. They also highlight how the framework captures diverse query characteristics, including comparison-style reasoning, tradeoff analysis, and open-ended decision-making. Together, these examples illustrate how the proposed framework produces structured and interpretable supervision signals for downstream reinforcement learning.
\vspace{0.5em}

\noindent
\begin{minipage}[t]{\columnwidth}
\centering

\scriptsize
\setlength{\tabcolsep}{3pt}
\renewcommand{\arraystretch}{1.08}

\vspace{0.4em}

\colorbox{gray!10}{
\parbox{0.96\linewidth}{
\vspace{2pt}
\centering
\normalsize
\textbf{Query:}
\textit{How does a vegetarian diet compare to a vegan one in terms of health and environmental impact?}
\vspace{2pt}
}}

\vspace{0.6em}

\begin{tabularx}{\linewidth}{|p{1.45cm}|p{1.15cm}|p{0.42cm}|X|}
\hline

\rowcolor{gray!15}
\textbf{Bucket} & \textbf{Category} & \textbf{Wt.} & \multicolumn{1}{c|}{\textbf{Rubric}} \\
\hline

\rowcolor{blue!2}
\multirow[t]{5}{1.45cm}{\textbf{Accuracy}}
& Essential & 5 & Explains the health benefits of both vegetarian and vegan diets, highlighting differences. \\
\rowcolor{blue!2}
& Essential & 5 & Compares environmental impacts including greenhouse gas emissions, water use, and land use. \\
\rowcolor{blue!2}
& Essential & 4 & Mentions that vegan diets generally have a lower environmental footprint than vegetarian diets. \\
\rowcolor{blue!2}
& Optional & 3 & Acknowledges that some plant-based foods can still have significant environmental impacts. \\
\rowcolor{blue!2}
& Essential & 4 & Includes specific evidence or findings from retrieved content to support environmental claims. \\
\hline

\rowcolor{orange!2}
\multirow[t]{3}{1.45cm}{\textbf{Composition}}
& Essential & 4 & Presents information directly without unnecessary restatement of the query. \\
\rowcolor{orange!2}
& Essential & 4 & Avoids unnecessary verbosity and provides a concise comparison. \\
\rowcolor{orange!2}
& Optional & 3 & Uses structured formatting to clearly compare health and environmental impacts when appropriate. \\
\hline

\rowcolor{green!2}
\multirow[t]{2}{1.45cm}{\textbf{Safety}}
& Essential & 5 & Avoids definitive health claims without acknowledging potential dietary deficiencies or risks. \\
\rowcolor{green!2}
& Essential & 5 & Handles dietary recommendations responsibly without prescriptive advice. \\
\hline

\rowcolor{purple!2}
\multirow[t]{3}{1.45cm}{\textbf{General}}
& Essential & 5 & Avoids unsupported claims about health impacts. \\
\rowcolor{purple!2}
& Optional & 3 & Includes relevant contextual caveats about environmental impacts of plant-based foods. \\
\rowcolor{purple!2}
& Essential & 5 & Avoids contradictions with retrieved evidence regarding environmental or health impacts. \\
\hline

\end{tabularx}

\end{minipage}

\vspace{1em}

\noindent
\begin{minipage}[t]{\columnwidth}
\centering

\scriptsize
\setlength{\tabcolsep}{3pt}
\renewcommand{\arraystretch}{1.08}

\vspace{0.4em}

\colorbox{gray!10}{
\parbox{0.96\linewidth}{
\vspace{2pt}
\centering
\normalsize
\textbf{Query:}
\textit{Should I buy an electric car now or wait a few years?}
\vspace{2pt}
}}

\vspace{0.6em}

\begin{tabularx}{\linewidth}{|p{1.45cm}|p{1.15cm}|p{0.42cm}|X|}
\hline

\rowcolor{gray!15}
\textbf{Bucket} & \textbf{Category} & \textbf{Wt.} & \multicolumn{1}{c|}{\textbf{Rubric}} \\
\hline

\rowcolor{blue!2}
\multirow[t]{5}{1.45cm}{\textbf{Accuracy}}
& Essential & 5 & Compares the advantages and disadvantages of buying an electric vehicle now versus waiting. \\
\rowcolor{blue!2}
& Essential & 5 & Discusses factors such as battery technology, charging infrastructure, and vehicle pricing. \\
\rowcolor{blue!2}
& Essential & 4 & Mentions financial considerations including incentives, depreciation, or long-term ownership costs. \\
\rowcolor{blue!2}
& Optional & 3 & Acknowledges that the best decision may depend on individual driving needs and geographic location. \\
\rowcolor{blue!2}
& Essential & 4 & Includes evidence-based reasoning or supporting trends when discussing future EV improvements. \\
\hline

\rowcolor{orange!2}
\multirow[t]{3}{1.45cm}{\textbf{Composition}}
& Essential & 4 & Organizes the comparison clearly across relevant decision factors. \\
\rowcolor{orange!2}
& Essential & 4 & Avoids unnecessary repetition or overly speculative discussion. \\
\rowcolor{orange!2}
& Optional & 3 & Uses concise formatting or structured comparison when appropriate. \\
\hline

\rowcolor{green!2}
\multirow[t]{2}{1.45cm}{\textbf{Safety}}
& Essential & 5 & Avoids definitive financial or purchasing advice without acknowledging uncertainty. \\
\rowcolor{green!2}
& Essential & 5 & Avoids unsupported claims about future EV technology or market conditions. \\
\hline

\rowcolor{purple!2}
\multirow[t]{3}{1.45cm}{\textbf{General}}
& Essential & 5 & Maintains consistency with retrieved evidence regarding EV adoption or infrastructure trends. \\
\rowcolor{purple!2}
& Optional & 3 & Includes contextual caveats about regional charging availability or policy incentives. \\
% \rowcolor{purple!2}
% & Essential & 5 & Avoids contradictory statements regarding the benefits or limitations of electric vehicles. \\
\hline

\end{tabularx}

\end{minipage}

\section{Qualitative Examples}
\label{app:qualitative-examples}

\vspace{0.5em}
We provide representative examples from Search Arena comparing outputs from the baseline and rubric-trained model variants, illustrating how grounding and multi-dimensional rubric design improve answer quality. Example~\ref{ex:fastapi-django} highlights improvements in factual grounding, and answer organization for a comparison-style query, while Example~\ref{ex:language} demonstrates improved synthesis, readability, and structural coherence for an open-ended recommendation query. Overall, the grounded variants provide stronger factual support, while the \structured\ Grounded model further improves response structure and overall answer quality.

\refstepcounter{example}
%\noindent\textbf{Example~\theexample. FastAPI vs Django}
\label{ex:fastapi-django}

\begin{examplebox}
\small

\textbf{Query}\\
which framework is better: FastAPI or Django, make your arguements

\vspace{4pt}
\centerline{\hdashrule[0.5ex]{\linewidth}{0.5pt}{3pt}}
\vspace{4pt}

\textbf{\textit{Baseline Answer}} \\
%\textbf{\textsc{Baseline Answer}} \\
Django is better for full-stack, complex web apps with built-in tools like ORM, admin interface, and authentication. FastAPI is better for fast, modern APIs and microservices, focusing on performance, simplicity, and modern Python features. [9]

\vspace{4pt}
\centerline{\hdashrule[0.5ex]{\linewidth}{0.5pt}{3pt}}
\vspace{4pt}

\textbf{\textit{Checklist-style Answer}} \\
Django is better for full-stack, complex web apps with built-in tools like ORM, admin interface, and authentication, making it suitable for rapid development and large-scale projects with extensive community support and resources. FastAPI, on the other hand, is ideal for fast, modern APIs and microservices, offering high performance, simplicity, and modern Python features like async capabilities and automatic documentation, making it a great choice for API-centric development and projects requiring high concurrency and low latency. Choose Django if you need a comprehensive framework with lots of built-in features and a strong emphasis on security, and opt for FastAPI if you prioritize lightweight, high-performance API development and modern Python features.

\vspace{4pt}
\centerline{\hdashrule[0.5ex]{\linewidth}{0.5pt}{3pt}}
\vspace{4pt}

\textbf{\textit{Single-Dimensional Grounded Answer}} \\
Django is better for full-stack, complex web apps with built-in tools like ORM, admin interface, and authentication, making it suitable for rapid development and large-scale projects. It has extensive community support and a robust framework with over 18 years of maturity [0][1][7][9][11].

FastAPI is more appropriate for fast, modern APIs and microservices, focusing on performance, simplicity, and modern Python features like async capabilities and automatic documentation. It's easier to learn and deploy, ideal for high-concurrency and API-centric development [0][1][5][8][9][13].

For a greenfield project, FastAPI is often recommended due to its speed and modern features, while Django is preferred for existing Django apps or projects requiring a full-stack solution [4][9].

\vspace{4pt}
\centerline{\hdashrule[0.5ex]{\linewidth}{0.5pt}{3pt}}
\vspace{4pt}

\textbf{\textit{Multi-Dimensional Grounded Answer}} \\
When deciding between FastAPI and Django, consider the following points:

\begin{itemize}
\item \textbf{Django:} A full-stack framework with extensive built-in features, strong community support, and an emphasis on security and rapid development. It is well-suited for complex web applications requiring components such as ORM, authentication, and admin interfaces [0][1][9][11].

\item \textbf{FastAPI:} A modern, high-performance framework optimized for APIs, leveraging Python type hints and asynchronous capabilities. It is particularly suitable for lightweight, high-concurrency, and API-centric applications [0][5][7][9].

\end{itemize}

Ultimately, the choice depends on project requirements, including the need for a full-stack solution versus API-focused development, performance considerations, and desired built-in functionality [0][1][5][7][9][11].

\end{examplebox}

\vspace{0.5em}
\noindent \textbf{Qualitative Analysis:}
The baseline response provides a broadly correct comparison but includes minimal supporting evidence and limited synthesis. The \checklist\ model improves coverage but remains verbose and lacks explicit grounding. The \flatrubrics\ Grounded model improves factual support by more consistently aligning claims with supporting citations and incorporating additional evidence-backed details. The \structured\ Grounded model further presents information in a more structured and coherent manner, improving readability, synthesis, and overall answer organization while maintaining strong grounding throughout the response.

\vspace{0.5em}
\refstepcounter{example}
\label{ex:language}

\begin{examplebox}
\small

\textbf{Query}\\
what is the best way to learn a new language

\vspace{2pt}
\centerline{\hdashrule[0.5ex]{\linewidth}{0.5pt}{3pt}}
\vspace{2pt}

\textbf{\textit{Baseline Answer}} \\
The most effective approach is usually a mix of consistent daily practice, lots of comprehensible input, and regular speaking and writing. Immersing yourself in listening and reading helps build understanding, while practicing output and interacting with others improves fluency. Using tools like apps can support learning, and making mistakes is an important part of the process. [5][2][1][6][11][0][18]

\vspace{2pt}
\centerline{\hdashrule[0.5ex]{\linewidth}{0.5pt}{3pt}}
\vspace{2pt}

\textbf{\textit{Checklist-style Answer}} \\
The best way to learn a new language involves consistent practice, immersion in the language environment, and using a combination of methods such as language apps, speaking with native speakers, and engaging with media in the target language [9]. Heavy input through reading and listening, noticing patterns, and gradually activating speaking and writing skills tend to outperform drills and formulas [2]. Immersion techniques like changing device settings to the target language, reading books and newspapers, and watching media without subtitles can mimic natural learning [3][10]. Connecting with native speakers, attending language schools, and participating in language exchanges are also effective [8][15]. Additionally, using flashcards, writing down new words, and making mistakes can enhance learning [11][16].

\vspace{2pt}
\centerline{\hdashrule[0.5ex]{\linewidth}{0.5pt}{3pt}}
\vspace{2pt}

\textbf{\textit{Single-Dimensional Grounded Answer}} \\
The best way to learn a new language involves consistent practice, immersion, and a combination of tools such as apps, speaking with native speakers, and engaging with media in the target language [9][2][3]. Heavy input through reading and listening, followed by gradual activation through speaking and writing, is particularly effective [2]. Immersion techniques like consuming media, reading content from native sources, and interacting with native speakers help reinforce learning [3][10][13]. Regular practice, making mistakes, and using structured methods such as flashcards further support fluency development [11][16].

\vspace{2pt}
\centerline{\hdashrule[0.5ex]{\linewidth}{0.5pt}{3pt}}
\vspace{2pt}

\textbf{\textit{Multi-Dimensional Grounded Answer}} \\
Learning a new language effectively involves several key strategies:
\begin{itemize}
\item \textbf{Heavy Input and Gradual Activation:} Focus on extensive reading and listening to build comprehension, and gradually move to speaking and writing ([2][0][5]).
\item \textbf{Consistent Practice and Immersion:} Regularly engage with the language through apps, media, and conversations with native speakers ([9][3][10]).
\item \textbf{Cultural Engagement:} Immerse yourself in the culture through TV shows, movies, and books to better understand context and nuances ([3][13]).
\item \textbf{Structured Learning:} Use formal instruction for grammar and pronunciation, supplemented by tools like apps and podcasts ([9][18]).
\item \textbf{Practice Conversations:} Engage in real or simulated conversations to build fluency and confidence ([5][1]).
\item \textbf{Personalized Learning:} Adapt methods to your interests to stay motivated, such as using flashcards or language exchanges ([16][17]).
\end{itemize}
Combining these strategies and maintaining consistency leads to effective language learning.

\end{examplebox}

\vspace{0.5em}
\noindent \textbf{Qualitative Analysis:}
The baseline response captures key ideas but remains brief and lacks organization. The \checklist\ model increases coverage but introduces redundancy and mixes related concepts without clear structure. The \flatrubrics\ Grounded model improves factual support and consistency but remains less structured in presentation. In contrast, the \structured\ Grounded model organizes the information into coherent categories, reduces redundancy, and integrates supporting evidence more effectively, resulting in a clearer, more structured, and more usable answer.

\vspace{8pt}

%\nolinenumbers

\section{Prompt Templates}
\label{app:prompts}
We share the prompt templates used throughout our framework, including rubric generation, rubric-based evaluation, grounded reference answer generation, and composition evaluation. Together, these prompts operationalize the rubric framework described in the main paper and support the various stages of training and evaluation.

\nolinenumbers
% \onecolumn

\begin{PromptBox}
\begin{PromptBlock}
# Objective

Given a user `Query`, `Retrieved Content` from a search engine, and a `Reference Answer`, generate structured rubrics organized into buckets that can be used to judge whether a response is helpful and high-quality.

A very important rule to follow is that all rubrics should be grounded in the `Retrieved Content` which is the actual content a system has access to when generating a response.

You will also be given reference answer(s). Treat them as **sample response(s)**, not as ground truth or strict constraints. Use it to:
- Understand the intent of the query and different aspects to cover
- Notice patterns of what a correct, and complete answer should contain

Overall, the rubrics should reflect what an ideal response must include, should include, or must avoid. They must be guided by the provided reference(s), but not limited to it. Use the reference(s) as a guideline, while ensuring the rubrics address the user query in its full complexity.

# Step 1: Classify the Query

Determine:
1. **Query category**: Identify the domain (e.g., technology, health, finance, travel, cooking, science, etc.). Output as: `<reason>...</reason> <domain>...</domain>`
2. **Time sensitivity**: Is a correct answer likely to change over time (e.g., current events, prices, software versions, live schedules)? Output YES or NO.

# Step 2: Generate Bucketed Evaluation Rubrics

Generate rubrics organized into the following four buckets. Each rubric must be ONE concise, testable sentence.

## Bucket: `accuracy`

Rubrics that assess whether the response correctly addresses the user’s query and includes the necessary information to produce a complete and correct answer.

Focus areas:
- Does the response correctly answer the question that is asked?
- Does the response directly answer the question, without missing the main intent or providing only partial or indirect answers?
- Does it correctly interpret the query intent (including noisy or speech-transcribed queries)?
- Does the response address all distinct aspects of the query, including multiple parts or constraints when present?
- Does the response include all essential facts, conditions, and required details needed to fully answer the query, as reflected in the Reference Answer?
- Does the response avoid introducing incorrect, unsupported, or contradictory information?
- Does the response stay focused on the query, avoiding irrelevant or off-topic content?
- When the query involves multiple plausible interpretations (e.g., entity, version, context), does the response clearly disambiguate or state its assumptions?

## Bucket: `composition`

Rubrics that assess the composition, structure, and presentation of the response content, focusing on how efficiently and directly the content is presented relative to the Reference Answer.

Focus areas:
- Does the response directly answer the user question, prioritizing need-to-know information first?
- Does the response avoid unnecessary verbosity or heavy formatting, and prioritize a simple, direct answer when appropriate?
- For simple queries, does the response prioritize conciseness and avoid unnecessary detail or elaboration?
- Does the response present information directly, without restating or interpreting the query (e.g., avoiding phrases like “the query asks” or “this question is about”), unless the query is genuinely ambiguous or underspecified?
- Does the response avoid describing the source or basis of the answer (e.g., “based on the provided passages/search results”, “based on the retrieved content”, “based on available information”), unless explicitly required by the query?
- Does the response avoid redundancy, including repeating or rephrasing the same information or splitting content into unnecessary sections?
- Does the response use structured formatting only when it improves clarity for multi-step tasks, comparisons, or atleast 3 discrete items, and avoid adding structure when a simple answer would suffice?
- Is the response appropriately concise for the query’s complexity — brief and direct for simple queries, while providing sufficient explanation for complex queries without unnecessary detail?
- If the query is ambiguous or underspecified, does the response handle ambiguity without extended meta-discussion?
- Does the response include only necessary caveats, and keep them concise rather than expanding into generic disclaimers?

## Bucket: `safety`

Rubrics that assess whether the response is safe and responsible.

Focus areas:
- Does the response avoid harmful, dangerous, or offensive content?
- Does the response handle sensitive topics (e.g., health, legal, financial) with appropriate caution?
- Does the response avoid unnecessarily alarming, fear-inducing, or extreme language, while still communicating relevant risks when appropriate?
- Does it refrain from giving definitive or prescriptive advice in domains requiring professional judgment?
- Does the response avoid providing actionable guidance that could lead to harm, especially in high-risk scenarios?

## Bucket: `general`

Rubrics that test for specific named facts, values, or mechanisms present in the Reference Answer — things a response could plausibly get wrong by omission or substitution.

Focus areas:
- Does the response rely on the provided context for its claims and avoid introducing information that cannot be supported or inferred from it?
- Does it avoid hallucinating facts that contradict the Reference Answer?
- Is the response free from internal contradictions or inconsistencies?
- Does the response avoid introducing specific facts that contradict the Reference Answer, even if they appear plausible?
- Does the response include necessary domain-specific details (e.g., parameters, units, versions, conditions) required to make the answer unambiguous and usable?
- When the answer depends on conditions, constraints, or prerequisites, does the response include those when necessary to avoid misleading or incomplete guidance?
- For time-sensitive or dynamic information (e.g., prices, schedules, availability), does the response provide appropriate temporal context (e.g., “as of” information) when necessary?

# Input

Query: {{ Query }}

Retrieved Content:
{{ Retrieved_Content }}

Reference Answer:
{{ Reference_Answer }}

# Output

Respond with a JSON object matching this schema exactly:

{
  "query_category": "<reason>...</reason> <domain>...</domain>",
  "time_sensitive": "YES",
  "rubrics_buckets": [
    {
      "id": "accuracy",
      "rubrics": [
        {
          "rubric": "The response directly answers the question asked.",
          "category": "Essential",
          "weight": 5
        }
      ]
    },
    {
      "id": "composition",
      "rubrics": []
    },
    {
      "id": "safety",
      "rubrics": []
    },
    {
      "id": "general",
      "rubrics": []
    }
  ]
}

All four bucket IDs (`accuracy`, `composition`, `safety`, `general`) must be present. Output only the JSON object. Do not include any additional text or explanation.

\end{PromptBlock}
\end{PromptBox}

\begin{center}
\captionof{figure}{Rubric Generation Prompt}
\label{fig:rubric-prompt}
\end{center}

% \begin{figure*}[t]
% \centering
% \begin{minipage}{0.98\textwidth}
\begin{PromptBox}
\begin{PromptBlock}
You are evaluating a response against specific rubric criteria.

Your role is to assess whether a generated response meets each rubric requirement. Each rubric belongs to a bucket and has a category (Essential or Optional) and weight.

## Instructions
- Evaluate based solely on what is present in the **Response** and **Query**.
- Do not introduce external facts or assumptions beyond the provided inputs.
- If the response contains "NO_ANSWER_FOUND", mark every rubric as unsatisfied (0).
- Accept equivalent phrasings or synonyms if they fulfill the rubric's intent.
- Ignore formatting, capitalization, or punctuation differences; focus on semantic meaning.
- For "must not" requirements, mark unsatisfied if the Response violates that constraint.
- If a Response both satisfies and violates parts of a rubric item, mark it as unsatisfied (0).
- Empty or irrelevant responses should fail all rubrics (0).
- Citations in square brackets like [1], [2,3], [4][5] are for reference linking and should not affect composition judgments.
- Maintain the original ordering of buckets and rubrics.

## Judging Criteria
- **Essential rubrics**: Mark satisfied (1) only if the Response fully meets all stated requirements and avoids all prohibited behaviors.
- **Optional rubrics**: Mark satisfied (1) if the Response includes the suggested element; otherwise unsatisfied (0).

## Output Format
Return JSON with the same structure as the input rubrics, adding only a `"satisfied"` field (0 or 1) to each rubric:

{
"buckets": [
  {
    "id": "<bucket id>",
    "rubrics": [
      {
        "rubric": "<exact rubric text>",
        "category": "<exact category>",
        "weight": <exact weight>,
        "satisfied": 0 or 1
      }
    ]
  }
]
}

## Requirements
- Copy "rubric", "category", and "weight" exactly as provided.
- Add "satisfied" as 0 (not satisfied) or 1 (satisfied).
- Do not add extra fields, comments, or explanations.
- Output valid JSON only.

--- INPUTS ---
Query: {{ query }}
Rubrics: {{ rubrics_json }}
Response: {{ response }}
\end{PromptBlock}
\end{PromptBox}
% \end{minipage}
% \captionsetup{justification=centering}
% \caption{Rubrics Judge Prompt}
% \label{fig:judge-prompt-1}
% \end{figure*}
\begin{center}
\captionof{figure}{Rubrics Judge Prompt}
\label{fig:judge-prompt-1}
\end{center}

% \begin{figure*}[t]
% \centering
% \begin{minipage}{0.98\textwidth}
\begin{PromptBox}
\begin{PromptBlock}
# Objective
Given a user `Query` and `Retrieved Content`, generate a concise, well-grounded answer that strictly relies on the provided content.

# Core Requirements

## 1. Grounding & Faithfulness
- Use ONLY information supported by the Retrieved Passages.
- Do NOT introduce external knowledge, assumptions, or hallucinations.
- Ignore irrelevant or redundant passages.

## 2. Citation (CRITICAL)
- Every factual statement MUST be supported by at least one citation.
- Use citation format: [index], where index corresponds to the passage.
- If a statement is supported by multiple passages, include all relevant citations (e.g., [2][5]).
- Do NOT include any statement without a citation.
- Ensure citations are accurate and directly support the claim they are attached to.
- Prefer more precise citations over broad or unnecessary ones.

## 3. Answer Style (Voice Assistant)
- Keep the response concise, natural, and easy to speak aloud.
- Avoid unnecessary verbosity, disclaimers, or meta commentary.
- Prioritize clarity and directness.

## 4. Handling Multiple Interpretations
- If the query has multiple plausible interpretations:
  - Start with a brief clarifying introduction.
  - Present each interpretation clearly as a short bullet or structured list.
  - EACH bullet MUST include supporting citations.
## 5. Noisy / Disfluent Queries
- The query may come from speech transcription and include errors.
- Infer intent using context from Retrieved Passages.
- Do NOT explicitly correct or comment on transcription errors.

## 6. General
- Do not mention retrieved content or passages anywhere in the answer.

# Input
Query: {{ Query }}

Retrieved Content:
{{ Retrieved_Content }}

# Output
Provide only the final answer following all rules above.

\end{PromptBlock}
\end{PromptBox}
% \end{minipage}
% \caption{Answer Generation Prompt}
% \label{fig:generation-prompt-1}
% \end{figure*}
\begin{center}
\captionof{figure}{Answer Generation Prompt}
\label{fig:generation-prompt-1}
\end{center}

\FloatBarrier
\begin{PromptBox}
\begin{PromptBlock}
# QA System Evaluation Prompt: Composition & Presentation

## Overview

This document provides a standardized evaluation rubric for assessing the **composition and presentation quality** of answers generated by a Question-Answering (QA) system. It is designed to be used by LLM-as-a-judge setups, human annotators, or both.

> **Note:** This rubric intentionally excludes factual accuracy. It focuses solely on *how* the answer is written, structured, and communicated.

---

## Evaluator Instructions

You are an expert evaluator assessing the quality of answers produced by a QA system. Your task is to evaluate each answer strictly on its **composition** and **presentation** — that is, how well the answer is written, structured, and communicated to the user, independent of factual correctness.

### Inputs
You will be given:
- **Query**: The user's original question.
- **Reference_Answer** (optional): A gold-standard or ideal answer for comparison.
- **Answer**: The answer generated by the QA system under evaluation.

## Evaluation Criteria

Rate the System Answer on each of the following **seven dimensions** using a **1–5 scale**.

### 1. Clarity (1–5)

- Is the answer easy to read and understand?
- Is the language precise and unambiguous?
- Are complex concepts explained in accessible terms when appropriate?

### 2. Conciseness (1–5)

- Does the answer avoid unnecessary verbosity, filler, or redundancy?
- Is every sentence purposeful and relevant to the question?
- Does it strike the right balance between thoroughness and brevity for the question's complexity?

### 3. Structure & Organization (1–5)

- Is the answer logically organized (e.g., introduction → body → conclusion, or step-by-step)?
- Does it use appropriate formatting aids (headings, bullet points, numbered lists, paragraphs) to enhance readability?
- Is there a coherent flow from one idea to the next?

### 4. Completeness of Response (1–5)

- Does the answer address all parts/sub-parts of the question?
- Are important caveats, edge cases, or nuances mentioned when relevant?
- Does the user need to ask follow-up questions to get a usable answer?

### 5. Tone & Register (1–5)

- Is the tone appropriate for the question's context (technical, conversational, formal, etc.)?
- Is the answer professional and neutral, avoiding unnecessary hedging or over-confidence?
- Does it match the apparent expertise level of the questioner?

### 6. Actionability & Usefulness (1–5)

- Can the user directly act on or apply the information provided?
- Are examples, code snippets, references, or next steps included where they would add value?
- Does the answer anticipate likely follow-up needs?

### 7. Formatting & Visual Presentation (1–5)

- Is whitespace used effectively to avoid walls of text?
- Are code blocks, tables, bold/italic emphasis, and other markdown elements used appropriately?
- Is the answer visually scannable — can a user quickly find the key information?

---

## Scoring Guide

| Score | Label        | Description                                                                 |
|-------|-------------|-----------------------------------------------------------------------------|
| 5     | Excellent   | Exceptionally well-composed; could serve as a gold-standard example.        |
| 4     | Good        | Well-composed with only minor areas for improvement.                        |
| 3     | Adequate    | Acceptable but with noticeable issues in composition or presentation.       |
| 2     | Below Average | Multiple composition/presentation problems that hinder understanding.     |
| 1     | Poor        | Poorly written, disorganized, or largely unusable in its current form.      |

---

## Output Format

For each evaluated answer, return your assessment in the following structured JSON format:

```json
{
    "clarity": {
      "score": "<1-5>",
      "justification": "<brief explanation>"
    },
    "conciseness": {
      "score": "<1-5>",
      "justification": "<brief explanation>"
    },
    "structure_and_organization": {
      "score": "<1-5>",
      "justification": "<brief explanation>"
    },
    "completeness_of_response": {
      "score": "<1-5>",
      "justification": "<brief explanation>"
    },
    "tone_and_register": {
      "score": "<1-5>",
      "justification": "<brief explanation>"
    },
    "actionability_and_usefulness": {
      "score": "<1-5>",
      "justification": "<brief explanation>"
    },
    "formatting_and_visual_presentation": {
      "score": "<1-5>",
      "justification": "<brief explanation>"
    },
    "overall_assessment": "<2-4 sentence summary of strengths and weaknesses>",
    "overall_score": "<weighted or simple average, rounded to 1 decimal>",
}
```

## Important Instructions

* Do NOT evaluate factual accuracy — focus solely on how the answer is composed and presented.
* Be calibrated — reserve scores of 5 for truly exceptional answers and scores of 1 for genuinely poor ones. Most answers should fall in the 2–4 range.
* Consider the question type — a simple factual question warrants a short, direct answer (don't penalize brevity); a complex how-to question warrants detailed structure (don't penalize length).
* Be specific in justifications — cite concrete phrases, structural choices, or formatting decisions rather than making generic statements.
* If a Reference_Answer is provided, use it as a benchmark for completeness and structure, but the Answer need not mirror it exactly to score well.

**Query**: {{Query}}
**Answer**: {{Answer}}

\end{PromptBlock}
\end{PromptBox}

\begin{center}
\captionof{figure}{Evaluation Prompt for Composition Axis}
\label{fig:composition-prompt}
\end{center}

% ======================== template from ACL =================================
\clearpage

\end{document}